\documentclass[runningheads]{llncs}
\usepackage[T1]{fontenc}
\usepackage{newtxtext}       %
\usepackage[varvw]{newtxmath}       

\usepackage{amsmath}
\usepackage{booktabs,multirow,array}
\usepackage[table]{xcolor}
\usepackage{tikz}
\usetikzlibrary{positioning,arrows.meta,calc}
\usepackage{cite}
\usepackage[
  colorlinks=true,
  citecolor=blue,
  urlcolor=blue,
  linkcolor=blue
]{hyperref}
\usepackage{algorithm}
\usepackage{algpseudocode}
\usepackage{enumitem}
\usepackage{placeins}
\usepackage[labelformat=parens,labelfont=bf,font=small]{subcaption}

\newcommand{\citep}[1]{\cite{#1}}

\definecolor{tsrsi}{HTML}{EEF4FB}
\definecolor{findingbg}{HTML}{F3F2EE}
\newcommand{\cyc}{CYC$^\ast$}
\newcommand{\cycval}{CYC$^\ast_{\mathrm{val}}$}
\newcommand{\rlin}{RLinear$^\ast$}
\newcommand{\tsc}{TS-RSI-Code}
\newcommand{\tss}{TS-RSI-Struct}

\usepackage{marvosym}

\usepackage{graphicx}
\begin{document}
\title{TS-RSI: Recursive Self-Improvement in Long-Term Time Series Forecasting}
%
%
%
%
\author{Xinhua Miao\inst{1} \and Bowei Yang\inst{2} \and Zhengong Cai\inst{1}\textsuperscript{\Letter}}
\authorrunning{X. Miao et al.}
%
\institute{School of Software Technology, Zhejiang University, Ningbo, China\\
\email{\{xinhuamiao,cstcaizg\}@zju.edu.cn}
\and
School of Aeronautics and Astronautics, Zhejiang University, Hangzhou, China\\
\email{boweiy@zju.edu.cn}}
\maketitle              
\begin{abstract}
Forecasting systems increasingly let LLMs write programs and revise them from validation feedback, a loop of recursive self-improvement (RSI). We introduce TS-RSI, a pre-registered testbed on long-term time series forecasting that pairs every loop with a trivial competitor encoding the same structure by hand. TS-RSI-Code uses one LLM-written program per dataset as a feature generator for a ridge correction to a closed-form CycleNet-style learner, with gap-matched fitting and a per-horizon fallback against drift. Relative to the CycleNet cycle, it lowers error by 14–27\% on the held-out PEMS datasets and costs at most 1\% on the other held-out datasets. The PEMS gain is a weekly cycle: choosing a daily or weekly cycle on validation does at least as well without an LLM, and iterating on the program improves the median by at most 0.8\% while validation overstates its gains. TS-RSI-Struct instead lets the LLM search the cycle length from validation feedback. Starting from a one-day prior on every dataset, it reaches the weekly cycle within three evaluations, beats random search by 4–7\% with three LLMs and ties the trivial competitor, for 0.24 CNY. When the prior is wrong on two structural choices, the loop falls behind random search.

\end{abstract}

\section{Introduction}
LLMs now write forecasting code that runs, and a growing body of work places them in loops that propose a program, score
it on validation data and keep the best candidates for the next
round~\citep{romeraparedes2024funsearch,novikov2025alphaevolve,lange2025shinka}. In time series, such loops write complete
forecasting pipelines~\citep{xu2026seats}, adapt the deployment of foundation models~\citep{liu2026timersi}, and learn
when to trust which expert~\citep{wang2026fase,hu2026timevolve}. Their gains are measured after many rounds of validation-based selection. When what the LLM contributes is a piece of structure, such as a seasonal profile, a model that is
handed that structure directly tells whether the loop adds anything.

\begin{figure*}[t]
\centering
\includegraphics[width=\linewidth]{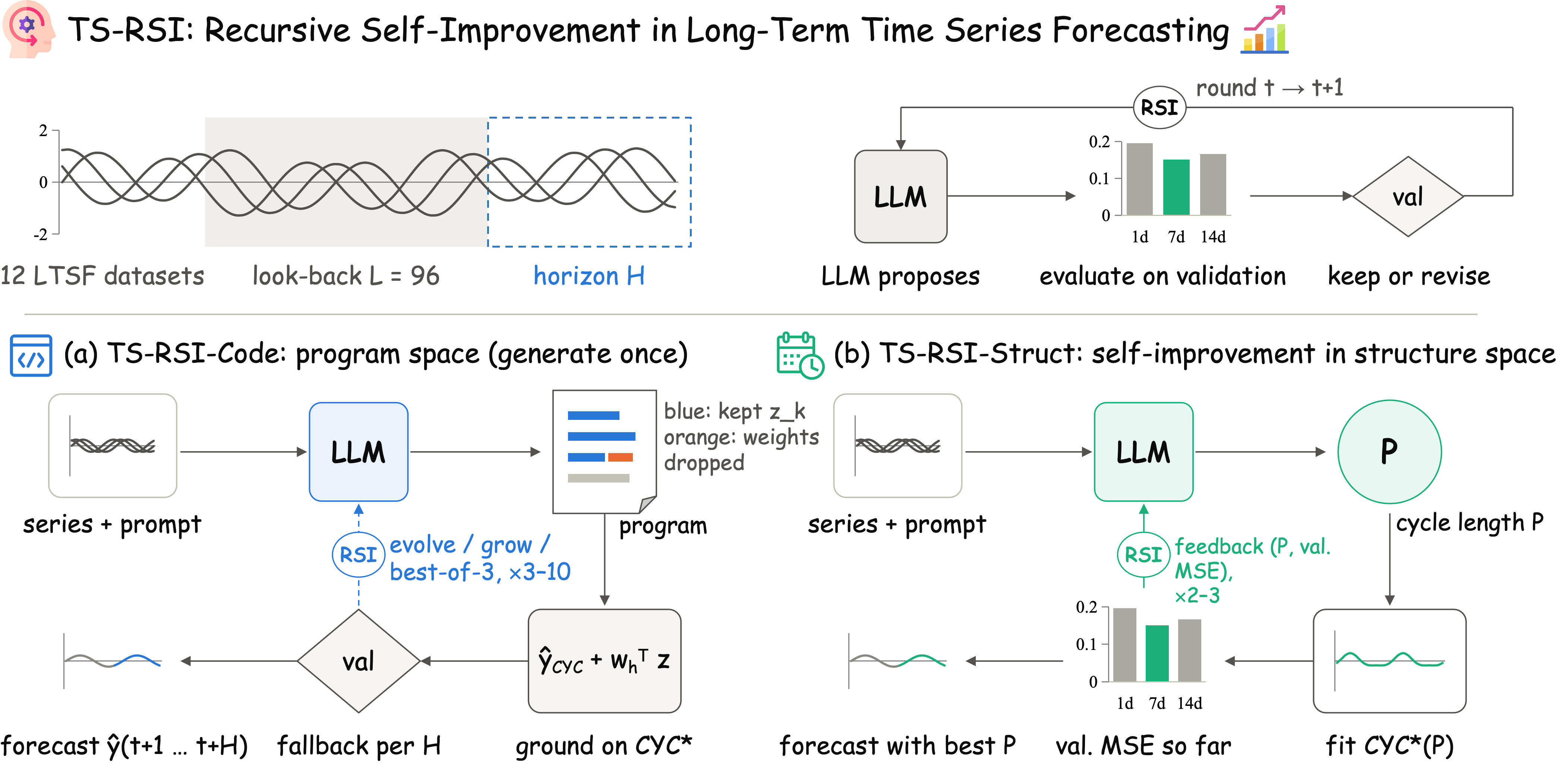}
\caption{Overview of TS-RSI. (a) \tsc: the LLM writes a program once, its arrays ($z_k$) become features of a ridge head on \cyc{}, and a
validation gate decides per horizon. Iterating (dashed) changes the median by about 1\%. (b) \tss: the LLM proposes the next cycle length $P$ from the validation MSE of those tried.}
\label{fig:overview}
\end{figure*}

TS-RSI measures this on the long-term forecasting (LTSF) benchmark on which most forecasting architectures are
compared~\citep{zhou2021informer,wu2021autoformer,zeng2023dlinear,nie2023patchtst,liu2024itransformer}. An LLM can
improve a forecaster in two places (Figure~\ref{fig:overview}): the code of a forecasting program, and a structural choice
of a closed-form model, here its cycle length. TS-RSI instantiates both. Code and supplementary materials for TS-RSI are available at \url{https://anonymous.4open.science/r/TS-RSI}. Our contributions are summarized as follows:

\begin{enumerate}[leftmargin=*]
\item \textbf{A testbed with trivial competitors.} Four development and eight held-out datasets, deterministic
closed-form learners on one CPU core, a ledger of every API call, and a pre-registration document for each of nine
experiments (Section~\ref{sec:tsrsi}).
\item \textbf{Program grounding (\tsc{}).} One LLM-written program per dataset serves as a feature generator for a ridge
correction to a CycleNet-style learner, with gap-matched fitting and a per-horizon fallback against drift. Relative to
the CycleNet cycle, it lowers error by 14--27\% on the held-out PEMS datasets and costs at most 1\% on the others
(Sections~\ref{sec:res-once} and~\ref{sec:res-drift}).
\item \textbf{What program search finds.} Five pre-registered ways of spending more calls and a ten-round loop improve the
median by at most 0.8\%, and validation overstates their gains (Section~\ref{sec:res-iter}). On PEMS the program
contributes a weekly cycle, which a two-way choice on validation also finds (Section~\ref{sec:res-trivial}).
\item \textbf{Structure-space self-improvement (\tss{}).} With validation feedback the LLM leaves its one-day prior,
matches exhaustive search, beats random search with three LLMs and ties the trivial competitor. When the prior is wrong
on two structural choices at once, it falls behind random search (Section~\ref{sec:res-struct}).
\end{enumerate}

\section{Related Work}
\subsection{LLM Program Search} FunSearch~\citep{romeraparedes2024funsearch}, AlphaEvolve~\citep{novikov2025alphaevolve}
and ShinkaEvolve~\citep{lange2025shinka} evolve programs with LLM mutations and an automatic evaluator. LLM-SR fits the
numeric parameters of LLM-written equation skeletons~\citep{shojaee2025llmsr}, and LLaMEA-HPO tunes the parameters of
LLM-designed heuristics inside the loop~\citep{vanstein2024llamea}. Our grounding step divides the labour the same way,
but fits weights on the internal variables of the program rather than on exposed parameters. CAAFE keeps LLM-written
tabular features that improve validation~\citep{hollmann2023caafe}. Our additive loop is its analogue for forecasting
programs. Gideoni et al.\ find simple baselines competitive with code evolution~\citep{gideoni2026simple}.

\subsection{LLMs as Optimizers and Hyperparameter Tuners} OPRO shows an LLM the solutions scored so far and asks for
better ones~\citep{yang2024opro}. The same loop tunes hyperparameters: the LLM proposes configurations from a task
description and the scores of earlier trials, and matches or beats random search and Bayesian optimization at small
budgets~\citep{zhang2023llmhpo} or the best human trials~\citep{liu2025agenthpo}. LLM-AutoOpt combines
Bayesian optimization with an LLM that reads dataset meta-features to tune time-series
forecasters~\citep{saadallah2026autoopt}. \tss{} is a loop of this kind over one structural choice.

\subsection{LLM-Based Forecasters} SEA-TS evolves forecasting code with a self-evolving agent~\citep{xu2026seats},
TimeRSI lets a coding agent revise the pipeline around a time-series foundation model (TSFM)~\citep{liu2026timersi}, and
ELATE evolves feature code~\citep{murray2025elate}. Online systems select or re-weight TSFMs from realised
errors~\citep{wang2026fase,hu2026timevolve} or evolve the guidance text of an LLM forecaster~\citep{mombaerts2026prompt}.
The latter reports, as we do, that most of the gain arrives in the first round of reflection. The LLM component of LLM-based forecasting models can often be
removed without loss~\citep{tan2024lmuseful}, and a recent survey reports similar ablations for agents~\citep{xu2026agentsurvey}.

\subsection{Selection Bias and Simple Baselines} Selecting among many candidates on one validation split biases the
selected score~\citep{cawley2010overfitting}. The reusable holdout limits the bias by releasing only coarse validation
information~\citep{dwork2015reusable}. The effect has been documented for research
agents~\citep{toledo2025aira,hambardzumyan2026aira2} and for keep-if-better self-improvement
loops~\citep{hu2026winners}. Time-series evaluation must respect temporal order and leave a gap between fitted and
evaluated data~\citep{bergmeir2012cv,lopezdeprado2018afml}. Our gap-matched grounding applies this to a stacked
correction~\citep{wolpert1992stacked}. Timestamp features help linear forecasters~\citep{ni2023mole,zeng2024timelinear}, and CycleNet~\citep{lin2024cyclenet} uses a daily cycle on PEMS and
Solar and a weekly cycle on ECL and Traffic. Our trivial competitor lets validation choose between the two.

\section{TS-RSI}\label{sec:tsrsi}
\subsection{Protocol}\label{sec:protocol}
\subsubsection{Data and splits.} We use the 12 datasets and splits of the Time-Series-Library protocol~\citep{wu2023timesnet}:
ETTh1, ETTh2, ETTm1, ETTm2~\citep{zhou2021informer}, Electricity (ECL), Traffic, Weather~\citep{wu2021autoformer},
Solar~\citep{lai2018lstnet} and PEMS03, PEMS04, PEMS07, PEMS08~\citep{liu2022scinet}, with look-back $L=96$, horizons
$\mathcal H=\{96,192,336,720\}$ ($\{12,24,48,96\}$ for PEMS), channels z-normalised with training statistics, and MSE on the normalised scale. ETTh1, Weather, ECL and PEMS04 serve to develop every component, and no component is fitted or
tuned on the other eight. All fitting uses the training split, and the validation split decides every selection and
fallback. Test scores of the program-space runs are computed after a run has finished. For the structure-space loops,
the validation and test scores of every candidate are computed once in advance, and the LLM sees only the validation
scores of the candidates it has tried.

\subsubsection{Pre-registration.} For each of the nine experiments on the held-out datasets we wrote a document that fixes
the method, the datasets, numeric success criteria with their interpretation, and MD5 prefixes of the code that would
run~\citep{nosek2018prereg}. The documents are not
entries in an external registry, and their header times were typed by hand: for three of them the header time is up
to eight minutes later than the first LLM call of the experiment in our API ledger. The held-out datasets were reused by
all nine experiments, and each experiment after the first was designed with earlier held-out results known.
Section~\ref{sec:limits} lists the cases that matter most. Analyses outside these documents are marked as exploratory.

\subsubsection{Deviations.} The weekly-cycle document required the trivial competitor as the reference of the main table,
the program re-grounded on its base as the main result, and no claim that \tsc{} improves on a CycleNet-style learner.
We keep \cyc{} and \tsc{} in Table~\ref{tab:main} because every program-space arm was pre-registered against them, add
the trivial competitor and the change relative to it, and state gains of \tsc{} only relative to the CycleNet cycle.
The re-grounded program fails its criterion (Section~\ref{sec:res-trivial}). No criterion was changed after its results
were known. In one document, the recorded hash of the generation script predates the prompt without component examples,
which was added before the first call. One script was patched and its run repeated after a network error had prevented
any LLM answer.

\subsubsection{LLM and compute.} Unless stated otherwise the LLM is DeepSeek-V4.1-Flash (non-thinking mode, temperature
0.7) behind an OpenAI-compatible API. DeepSeek-V4-Pro serves as the stronger LLM and Qwen3.6-Plus as a second model
family. All learners are closed form, deterministic and run on one CPU core without a GPU, so run-to-run variation comes
only from LLM sampling.

\subsection{Closed-Form Bases}\label{sec:bases}
Both instantiations of TS-RSI improve a closed-form forecaster. For a window ending at time $t$, a forecaster maps
$x_{t-L+1:t}\in\mathbb R^{L\times C}$ to $\hat y_{t+1:t+H}\in\mathbb R^{H\times C}$.
\rlin{} is RLinear~\citep{li2023rlinear}: a channel-shared linear map, fitted by ridge regression on
instance-normalised windows. \cyc{} is a closed-form analogue of CycleNet/Linear~\citep{lin2024cyclenet}. For a cycle
length $P$ it estimates, for every channel $c$, the mean of the training series at each phase,
\begin{equation}
Q_c(\phi)=\frac{1}{|\mathcal T_\phi|}\sum_{\tau\in\mathcal T_\phi}x_{\tau,c},\qquad
\mathcal T_\phi=\{\tau\le T_{\mathrm{train}}:\tau \bmod P=\phi\},
\label{eq:profile}
\end{equation}
removes it from the window, $r_\tau=x_\tau-Q(\tau \bmod P)$, normalises the remainder with its window mean $\mu$ and
standard deviation $\sigma$, and forecasts with a ridge-fitted map $W_h\in\mathbb R^{L+1}$ shared across channels:
\begin{equation}
\hat y^{\mathrm{CYC}}_{t+h}=\sigma\,W_h^{\top}\!\left[\frac{r_{t-L+1:t}-\mu}{\sigma};\,1\right]+\mu+Q\big((t+h)\bmod P\big),
\qquad h=1,\dots,H.
\label{eq:cyc}
\end{equation}
By default $P$ follows CycleNet (24 for ETTh, 96 for ETTm, 144 for Weather and Solar, 168 for ECL and Traffic, 288 for
PEMS). The cycle length $P$ is the structural choice that \tss{} searches.

\subsection{TS-RSI-Code: Program Space}\label{sec:method}
\subsubsection{Generation.} The LLM receives a short description of the dataset and an interface: a class with
\texttt{fit(train)} and a vectorised \texttt{predict(x, t\_start)}, where \texttt{t\_start} is the absolute time index of
the first forecast step. The prompt asks for an explicit combination of components, each in its own variable, and lists
examples such as level, damped trend, seasonal naive, daily and weekly profiles and cross-channel factors. A program
$\pi$ is accepted after a smoke test and a scale check: fitted on the first 80\% of the training split, its MSE on the
last 20\% must not exceed three times that of repeating the last value. A failing program is returned to the LLM for repair.

\subsubsection{Grounding.} A generated program can compute useful components even when its hand-set weights fit the data
poorly, so we use it as a feature generator and learn the weights. Let $E$ be the end of the fitted rows (the end of the
training split) and $b=\max(0.2E,\,H+128)$. The base and the program are first fitted on the rows before $E-b$, and the head
is fitted on the windows $\mathcal W$ whose targets lie in $[E-b,E)$, which are out of sample for both (at most 256
windows and 32 channels). The \texttt{predict} method of $\pi$ is rewritten to record its forecast and every assigned
array whose shape aligns with the forecast. Arrays that are constant across windows or nearly duplicate another array
(correlation above 0.999) are dropped, and each remaining intermediate is kept only if perturbing it changes the
forecast of $\pi$. This leaves $K$ features $a_k\in\mathbb R^{H\times C}$. Each is clipped to the 2nd and 98th
percentiles $[l_k,u_k]$ of its values on $\mathcal W$, so that a drifting intermediate cannot extrapolate the correction,
and standardised with its mean $m_k$ and standard deviation $s_k$ on $\mathcal W$,
$\tilde z_k=(\mathrm{clip}(a_k,l_k,u_k)-m_k)/s_k$. For every horizon step $h$ the grounded forecast corrects the base,
\begin{equation}
\hat y^{\mathrm{G}}_{t+h}=\hat y^{\mathrm{CYC}}_{t+h}+\beta_h^{\top}\zeta_{t+h},\qquad
\zeta_{t+h}=\big[\hat y^{\mathrm{CYC}}_{t+h},\ \tilde z_{1,t+h},\dots,\tilde z_{K,t+h},\ 1\big],
\label{eq:ground}
\end{equation}
where $\beta_h$ is the solution of a ridge regression of the base residual, pooled over channels and windows,
\begin{equation}
\beta_h=\arg\min_{\beta}\sum_{n\in\mathcal W}\sum_{c}\Big(y_{n,h,c}-\hat y^{\mathrm{CYC}}_{n,h,c}-\beta^{\top}\zeta_{n,h,c}\Big)^2+\lambda_h\|\beta_{1:K+1}\|^2 .
\label{eq:ridge}
\end{equation}
The intercept is not penalised, $\lambda_h=\lambda_{\mathrm{rel}}\cdot\overline{\mathrm{diag}}(\sum\zeta\zeta^{\top})$, and
$\lambda_{\mathrm{rel}}$ is chosen on a chronological 70/30 split of $\mathcal W$. For deployment the base and the program
are refitted on all rows before $E$, keeping $\beta_h$ and the clipping and standardisation constants.

\subsubsection{Robust grounding.} The test split begins $n_{\mathrm{val}}$ rows (one validation length) after the fitted rows, while the windows
$\mathcal W$ directly follow the rows the head-fit models saw. In \emph{gap-matched fitting} these models are fitted on
the rows before \mbox{$E-b-n_{\mathrm{val}}$} instead of $E-b$, so that $\mathcal W$ sees the same distance between fitted rows and evaluated
windows as the test split. Deployment again uses the models refitted up to $E$. In the \emph{per-horizon fallback}, for
each forecast length $H$ the grounded forecast is used only if it beats the base on validation:
\begin{equation}
\hat y^{\mathrm{Code}}_{t+1:t+H}=\begin{cases}\hat y^{\mathrm{G}}_{t+1:t+H}&\text{if }\mathrm{MSE}_{\mathrm{val}}(\hat y^{\mathrm{G}};H)<\mathrm{MSE}_{\mathrm{val}}(\hat y^{\mathrm{CYC}};H),\\[2pt]
\hat y^{\mathrm{CYC}}_{t+1:t+H}&\text{otherwise.}\end{cases}
\label{eq:fallback}
\end{equation}
\tsc{} uses both (Section~\ref{sec:res-drift} explains why). Grounding on \cyc{} without them is called CLG and serves as an ablation.

\subsubsection{Iterating in program space.} Starting from \tsc{}, we pre-registered five ways of spending more calls, all
scored with the same grounding and fallback. The additive loop \emph{grow} adds programs to a pool whose features share
one head: in each of three rounds the LLM sees residual diagnostics on a feedback fold inside the training split (error by
horizon, phase of day, day of week, cross-channel structure) and writes two complementary programs, and the better one is
kept only if both the feedback fold and validation improve by more than 2\%. (The grow and evolve prompts describe the
base as a fixed linear model and do not mention the periodic profile of \cyc{}.) The rewrite loop \emph{evolve} rewrites the current best program after seeing validation diagnostics and keeps the candidate with the best
validation MSE (three rounds of two candidates). \emph{Best-of-3} and \emph{union-3} draw three independent programs and
select one on validation or pool their features. The \emph{stronger-LLM} arm generates once with DeepSeek-V4-Pro.
A later experiment runs evolve for ten rounds.

\subsection{TS-RSI-Struct: Structure Space}\label{sec:structure}
The structure we vary is the cycle length of \cyc{}, drawn from six candidates per dataset, $\mathcal P=\{1,\,d/2,\,d,\,2d,\,7d,\,14d\}$
steps, where $d$ is the number of steps per day (no cycle, 12 hours, and 1, 2, 7 and 14 days). A candidate is scored by the
validation MSE of \cyc{} with that cycle length, averaged over the four horizons,
\begin{equation}
V(P)=\frac{1}{|\mathcal H|}\sum_{H\in\mathcal H}\mathrm{MSE}_{\mathrm{val}}\big(\hat y^{\mathrm{CYC}(P)};H\big).
\label{eq:val}
\end{equation}
\emph{\tss{}}, an LLM-as-optimizer loop~\citep{yang2024opro,zhang2023llmhpo}, proposes a first candidate from the dataset
description $\mathcal D$ alone and then, in each round, sees the candidates tried so far with their scores and proposes an untried one,
\begin{equation}
P_1\sim\pi_{\mathrm{LLM}}(\mathcal D),\qquad
P_{j+1}\sim\pi_{\mathrm{LLM}}\big(\mathcal D,\{(P_i,V(P_i))\}_{i\le j}\big),\quad P_{j+1}\in\mathcal P\setminus\{P_1,\dots,P_j\},
\label{eq:struct}
\end{equation}
and deploys the best candidate on validation after $k$ evaluations ($k-1$ rounds of feedback),
$\hat P_k=\arg\min_{i\le k}V(P_i)$. The LLM never sees test data. We run three seeds per
dataset and deploy $k=3$ unless stated. An invalid or repeated answer is replaced by the first untried candidate.

\subsubsection{Reference rules.} Three rules choose the cycle length without iterating on test data. \emph{Default} uses the
CycleNet value. The \emph{trivial competitor} \cycval{} picks the daily or the weekly cycle by validation,
$\hat P=\arg\min_{P\in\{d,7d\}}V(P)$, without an LLM. The \emph{LLM prior} ranks the six candidates from the dataset
description alone (three samples, Borda consensus). Three reference searches complete the comparison: six-way validation
search, $\arg\min_{P\in\mathcal P}V(P)$; random search over $k$ candidates, whose test MSE is the expectation over all
$\binom{6}{k}$ subsets; and the LLM's static ranking truncated at $k$.

\subsubsection{Aggregating over datasets.} We report the change of a method $a$ relative to a method $b$ as the geometric
mean of $\mathrm{MSE}_s(a)/\mathrm{MSE}_s(b)$ over a set of datasets $s\in\mathcal S$, minus one, and use the median of
the per-dataset ratios where a criterion says so.

\section{Experimental Results}
\subsection{One Generated Program Helps Where Seasonality Matters}\label{sec:res-once}
\begin{table*}[t]
\centering
\caption{Test MSE averaged over four horizons (top: development datasets). Naive: last value; SN$_{\mathrm{val}}$:
seasonal naive with a daily or weekly period picked on validation (both exploratory). Published numbers
from~\citep{lin2024cyclenet,liu2024itransformer} (CN: CycleNet). \cyc{}: CycleNet cycle; \cycval{}: trivial
competitor; Code: first generation, mean of three in parentheses; Struct: after three evaluations, mean of three
seeds. Bold/underlined: best/second.}
\label{tab:main}
\scalebox{0.678}{\begin{tabular}{l|cc|cccccc|c>{\columncolor{tsrsi}}c>{\columncolor{tsrsi}}cc|>{\columncolor{tsrsi}}c>{\columncolor{tsrsi}}c|>{\columncolor{tsrsi}}c>{\columncolor{tsrsi}}c}
\toprule
\multirow{2}{*}{Dataset} & \multicolumn{2}{c|}{Naive} & \multicolumn{6}{c|}{Published} & \multicolumn{4}{c|}{Ours (Closed Form)} & \multicolumn{2}{c|}{vs.\ \cyc{} (\%)} & \multicolumn{2}{c}{vs.\ \cycval{} (\%)} \\ \cmidrule{2-17}
 & Naive & SN$_{\mathrm{val}}$ & DLinear & PatchTST & iTrans. & RLinear & CN/Lin. & CN/MLP & \cyc{} & \textbf{Code} & \textbf{Struct} & \cycval{} & \textbf{Code} & \textbf{Struct} & \textbf{Code} & \textbf{Struct} \\
\midrule
ETTh1 & 1.321 & 0.600 & 0.456 & 0.469 & 0.454 & 0.446 & \underline{0.432} & 0.457 & \textbf{0.424} & \textbf{0.424}{\tiny\,(.549)} & 0.439 & 0.436 & $0.0$ & $+3.5$ & $-2.8$ & $+0.6$ \\
Weather & 0.352 & 0.371 & 0.265 & 0.259 & 0.258 & 0.272 & \underline{0.254} & \textbf{0.243} & 0.293 & 0.340{\tiny\,(.352)} & 0.293 & 0.293 & $+16.0$ & $0.0$ & $+16.0$ & $0.0$ \\
ECL & 1.612 & 0.231 & 0.212 & 0.205 & 0.178 & 0.219 & 0.170 & \underline{0.168} & 0.170 & \textbf{0.167}{\tiny\,(.167)} & 0.170 & 0.170 & $-1.8$ & $0.0$ & $-1.8$ & $0.0$ \\
PEMS04 & 0.623 & 0.200 & 0.295 & 0.195 & 0.111 & 0.526 & 0.144 & 0.119 & 0.154 & \underline{0.101}{\tiny\,(.117)} & \textbf{0.097} & \textbf{0.097} & $-34.3$ & $-36.8$ & $+3.8$ & $0.0$ \\
\midrule
ETTh2 & 0.539 & 0.483 & 0.559 & 0.387 & 0.383 & \underline{0.374} & 0.383 & 0.388 & \textbf{0.370} & \textbf{0.370}{\tiny\,(.373)} & \textbf{0.370} & \textbf{0.370} & $0.0$ & $0.0$ & $0.0$ & $0.0$ \\
ETTm1 & 1.271 & 0.489 & 0.403 & 0.387 & 0.407 & 0.414 & \underline{0.386} & \textbf{0.379} & \underline{0.386} & \underline{0.386}{\tiny\,(.386)} & \underline{0.386} & \underline{0.386} & $-0.1$ & $0.0$ & $-0.1$ & $0.0$ \\
ETTm2 & 0.385 & 0.358 & 0.350 & 0.281 & 0.288 & 0.286 & \underline{0.272} & \textbf{0.266} & 0.273 & 0.274{\tiny\,(.274)} & 0.273 & 0.273 & $+0.2$ & $0.0$ & $+0.2$ & $0.0$ \\
Traffic & 2.765 & 0.634 & 0.625 & 0.481 & \textbf{0.428} & 0.626 & 0.485 & \underline{0.472} & 0.486 & 0.483{\tiny\,(.485)} & 0.486 & 0.486 & $-0.6$ & $0.0$ & $-0.6$ & $0.0$ \\
Solar & 1.451 & 0.341 & 0.330 & 0.270 & \underline{0.233} & 0.369 & 0.235 & \textbf{0.210} & 0.248 & \underline{0.233}{\tiny\,(.234)} & 0.250 & 0.248 & $-5.8$ & $+1.1$ & $-5.8$ & $+1.1$ \\
PEMS03 & 0.572 & 0.275 & 0.278 & 0.180 & \textbf{0.113} & 0.495 & 0.139 & \underline{0.118} & 0.146 & 0.126{\tiny\,(.124)} & 0.122 & 0.122 & $-13.7$ & $-16.4$ & $+3.2$ & $0.0$ \\
PEMS07 & 0.603 & 0.169 & 0.329 & 0.211 & \underline{0.101} & 0.505 & 0.138 & 0.113 & 0.141 & 0.103{\tiny\,(.112)} & \textbf{0.081} & \textbf{0.081} & $-26.8$ & $-42.5$ & $+27.4$ & $0.0$ \\
PEMS08 & 0.641 & 0.302 & 0.379 & 0.280 & \underline{0.150} & 0.529 & 0.176 & \underline{0.150} & 0.187 & 0.153{\tiny\,(.159)} & \textbf{0.134} & \textbf{0.134} & $-18.3$ & $-28.5$ & $+14.2$ & $0.0$ \\
\bottomrule
\end{tabular}}
\end{table*}

The closed-form bases track their published counterparts: \rlin{} is within $\pm4.2\%$ of RLinear on all 12 datasets,
and \cyc{} is within $\pm1\%$ of CycleNet/Linear on ETTm1, ETTm2, ECL and Traffic, better on ETTh1 and ETTh2, and worse
on Solar ($+5\%$), PEMS ($+2$ to $+7\%$) and Weather ($+15\%$).

On the held-out datasets all three pre-registered criteria for \tsc{} against \cyc{} hold (Table~\ref{tab:main}). It is
within 1\% of \cyc{} on 8 of 8 datasets, against 6 of 8 for plain CLG, which loses 12\% on ETTh2 and 6\% on ETTm2; its
median change is $-3.2\%$; and it keeps a median 99\% of the gain wherever CLG gains at least 5\%. The gains come from
calendar-driven data: $-13.7\%$ on PEMS03, $-26.8\%$ on PEMS07, $-18.3\%$ on PEMS08 and $-5.8\%$ on Solar. The published models stay ahead on ETTm1, ETTm2, Traffic,
Solar and Weather, and iTransformer on PEMS03 (0.113 against 0.122).

Table~\ref{tab:main} reports the program generated first, which was fixed before the other two. Over three independent
generations the held-out median change of \tsc{} is $-3.2\%$, $-3.2\%$ and $-2.5\%$, and every generation stays within
1\% of \cyc{} on at least 7 of the 8 held-out datasets. Averaged over the three generations, the PEMS gains are 15.0\%,
20.3\% and 14.9\%. The spread is large only on PEMS07 ($-29.8\%$ to $-4.5\%$) and on the development dataset ETTh1, where
the first program is the one gap-matched fitting was developed on (Section~\ref{sec:res-drift}).

\begin{figure*}[t]
\centering
\includegraphics[width=\linewidth]{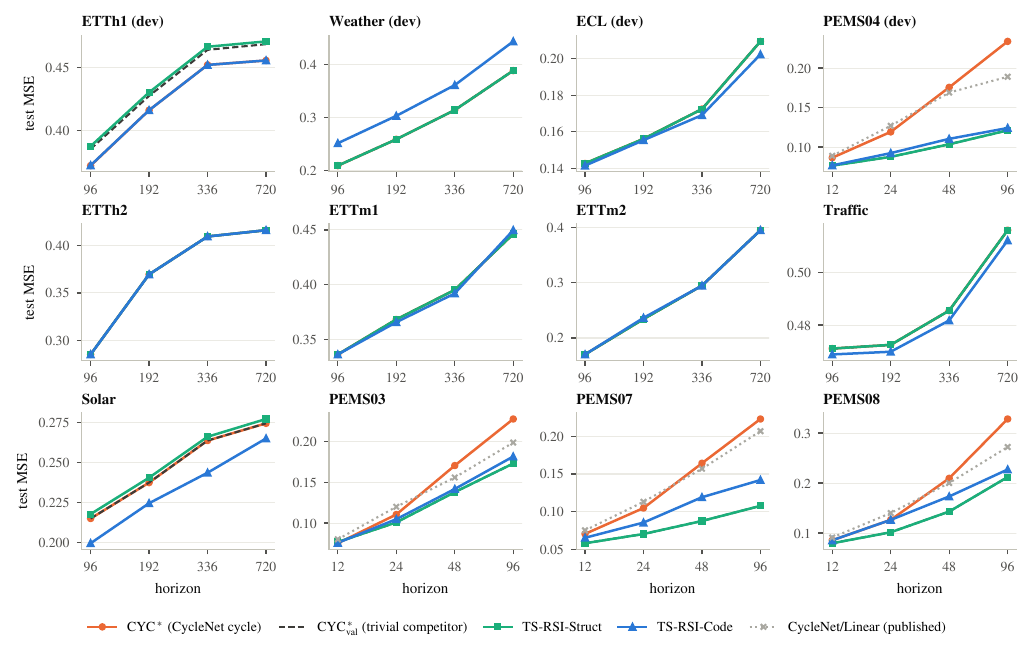}
\caption{Test MSE per horizon. On PEMS the gain of \tsc{} and \tss{} over \cyc{} grows with the horizon, and \tss{}
coincides with \cycval{}.}
\label{fig:horizon}
\end{figure*}

The gain grows with the horizon (Figure~\ref{fig:horizon}). On PEMS07
\tsc{} is 7.3\% better than \cyc{} at 12 steps and 36.4\% better at 96 steps. On PEMS08 the change goes from 0.0\% to
$-30.8\%$ and on PEMS03 from $+0.7\%$ to $-20.2\%$. A seasonal profile matters little for the next hour, which the
look-back window already describes, and a great deal for the next eight hours.

\begin{figure*}[t]
\centering
\includegraphics[width=\linewidth]{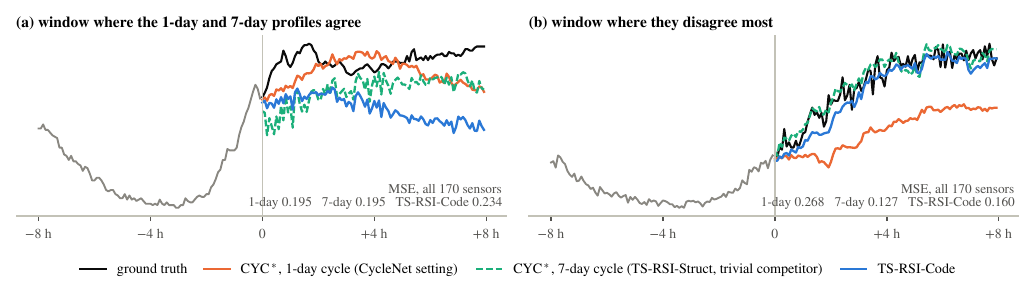}
\caption{PEMS08 forecasts from 06:00 where the daily and weekly profiles of \cyc{} agree most (a) or disagree most (b),
for the sensor where they differ most.}
\label{fig:showcase}
\end{figure*}

Figure~\ref{fig:showcase} shows two PEMS08 test windows: where the daily and weekly profiles disagree, \tsc{} stays
close to the weekly one. Where they agree, it is worse than both, and the per-horizon fallback, which decides on
averages over the validation split, cannot avoid such windows.

\subsection{Grounding Under Drift}\label{sec:res-drift}
\begin{figure*}[t]
\centering
\includegraphics[width=\linewidth]{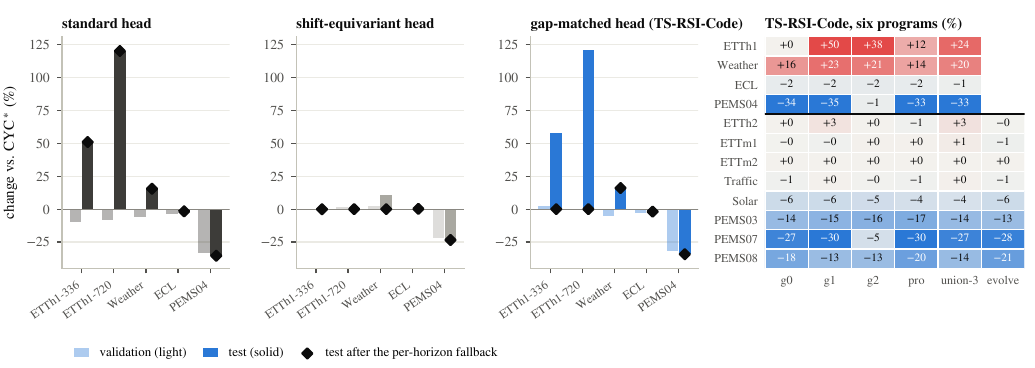}
\caption{Left: change of CLG relative to \cyc{} on the development datasets for three correction heads. Right: \tsc{}
relative to \cyc{} in \% (top: development, bottom: held out) for generations g0--g2, the
stronger LLM (pro), union-3 and evolve.}
\label{fig:drift}
\end{figure*}

Two choices in \tsc{} come from a failure on the development datasets. CLG improved validation by 6\% on ETTh1 and
Weather and lost 46\% and 15\% on test. Figure~\ref{fig:drift} (left) shows the screen that located the cause. A head
that is exactly shift-equivariant, with no term that anchors the forecast to the training-period level, removes the
ETTh1 loss but also all of the ECL gain and a third of the PEMS04 gain. Regressing toward the training-period level helps
on stationary data and hurts under drift, and validation, which directly follows the training data, cannot tell the two
apart. Fitting the head at the deployment gap lets validation see the drift: on ETTh1 at $H=336$ the validation change
turns from $-9.6\%$ to $+2.6\%$, the fallback switches to the base, and the ECL and PEMS04 gains are unchanged. With the fallback, the shift-equivariant head would also have protected Weather, where the gap-matched head loses 16\%.

The fix is partial (Figure~\ref{fig:drift}, right). On the held-out datasets each single program stays within 1\% of the
base on at least 7 of 8 datasets, and the union of three programs on 6 of 8. On ETTh1 the fix protects only the program
it was developed on: with the other programs, validation still improves at long horizons while test loses 12--50\%.
Weather loses 14--23\% with every program.

\subsection{Iterating on the Program Adds Little}\label{sec:res-iter}
\begin{figure*}[t]
\centering
\includegraphics[width=\linewidth]{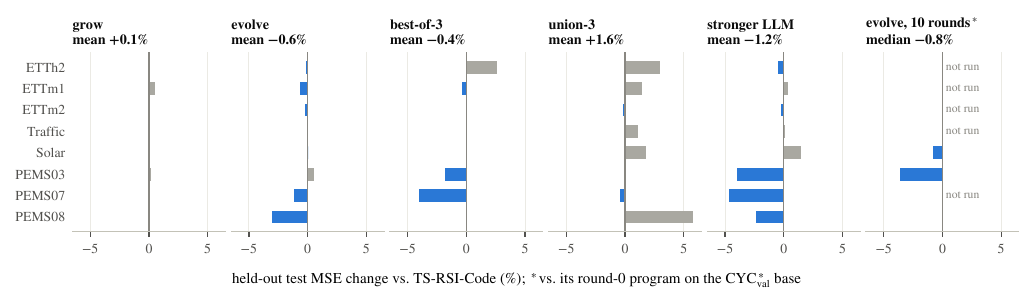}
\caption{Test MSE change of each program-space arm relative to \tsc{}, per held-out dataset (negative is better). The
last panel is relative to its round-0 program.}
\label{fig:arms}
\end{figure*}

Spending more calls on the program changes little (Figure~\ref{fig:arms}). None of the five pre-registered arms improves
the held-out median by more than 0.4\%. In the mean, only the stronger LLM gains more than 1\% ($-1.2\%$), and union-3
loses 1.6\%. The additive loop accepted a program on only two datasets, and both became slightly worse on test. A multi-fold version of the rewrite loop, run earlier on the
development datasets, failed both of its criteria (better on 2 of 4, mean $+5.7\%$).

\begin{figure}[t]
\centering
\begin{subfigure}[t]{0.4\linewidth}\centering
\includegraphics[width=\linewidth]{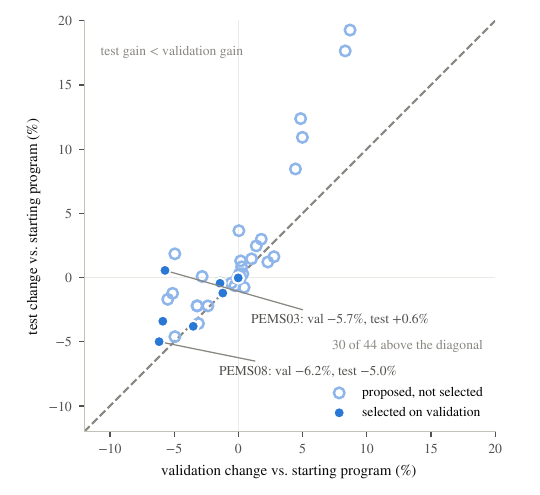}
\caption{All 44 runnable candidates.}
\end{subfigure}\hfill
\begin{subfigure}[t]{0.6\linewidth}\centering
\includegraphics[width=\linewidth]{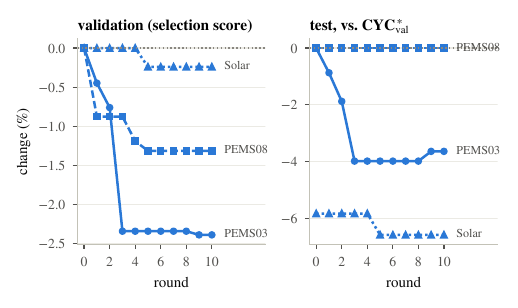}
\caption{Ten rounds of evolve, best program so far.}
\end{subfigure}
\caption{Validation overstates the gains of iterated programs. (a) Validation vs.\ test change of evolve's candidates
relative to the starting program (before fallback), with selected ones filled. Points above the diagonal gained less on test
than on validation. (b) Best so far over ten rounds: validation relative to round 0, test (with fallback) relative to
\cycval{}.}
\label{fig:curse}
\end{figure}

\subsubsection{Validation overstates gains.} The pre-registered criterion holds: on 5 of the 7 held-out datasets where the
rewrite loop replaced the starting program, the selected program gained less on test than on validation, by a median of
1.0 percentage points. Over all 44 runnable programs the loop proposed
(exploratory), validation ranks them well (rank correlation 0.79), and 30 of them did better on validation than on test
(Figure~\ref{fig:curse}a).

\subsubsection{A larger budget.} To test whether the loops are simply too short, we pre-registered ten rounds of evolve on
PEMS03, PEMS08 and Solar, chosen after the results of Section~\ref{sec:res-trivial} were known: the trivial competitor
picks the weekly cycle on the first two, and Solar is where \tsc{} kept a gain over it. The loop starts from the \tsc{}
program re-grounded on the base chosen by the trivial competitor, so any gain is a gain over that competitor
(Figure~\ref{fig:curse}b). Against its round-0 program the median
change is $-0.8\%$, short of the pre-registered $-1\%$. Against the trivial competitor it is $-3.6\%$, which passes the
second criterion. On Solar, $-5.8\%$ of the $-6.6\%$ is already in the round-0 program. Without the fallback, validation gains ($-2.4\%$, $-1.3\%$, $-0.2\%$) again differ from test gains
($-1.0\%$, $-0.5\%$, $-0.8\%$).

\subsection{A Trivial Competitor Explains the Gain}\label{sec:res-trivial}
\begin{figure}[t]
\centering
\begin{subfigure}[t]{0.495\linewidth}\centering
\includegraphics[width=\linewidth]{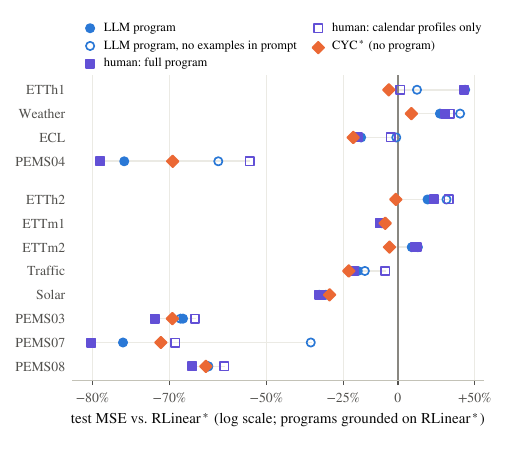}
\caption{Programs grounded on \rlin{}.}
\end{subfigure}\hfill
\begin{subfigure}[t]{0.495\linewidth}\centering
\includegraphics[width=\linewidth]{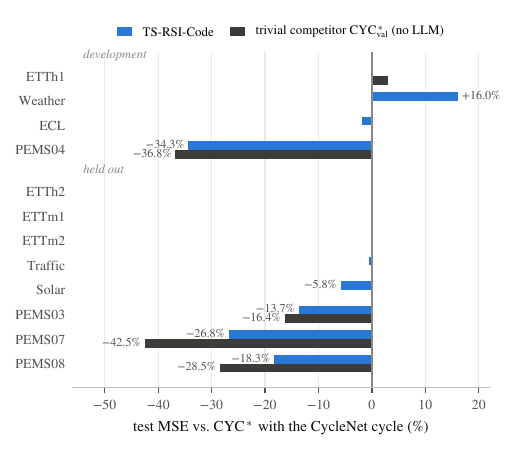}
\caption{\tsc{} vs.\ trivial competitor.}
\end{subfigure}
\caption{Where the gain comes from.}
\label{fig:attr}
\end{figure}

Grounded the same way on \rlin, the LLM program beats a human-written calendar program (daily and weekly profiles) on 9
of 12 datasets with a median of $-7.1\%$, is slightly worse than a human answer to the same prompt (better on 3 of 12,
median $+1.8\%$), and beats \cyc{} on only 4 of 12 (Figure~\ref{fig:attr}a). Without component examples in the prompt,
36 of 36 generations still succeed; their held-out median is $-25.0\%$ against \rlin, versus $-26.1\%$ with examples.
What the program contributes is mostly periodic structure.

All twelve generated programs compute a time-of-week profile, and so do 9 of 12 programs generated for PEMS without
examples. The large gains appear on PEMS, where \cyc{} uses a daily cycle for traffic that follows a weekly one. After the
held-out results of \tsc{} were known, we therefore pre-registered the trivial competitor \cycval{} with a criterion: if
the median change of \tsc{} against \cycval{} on the held-out datasets is above $-1\%$, the weekly cycle explains the
gain of the LLM program.

The median is $+0.1\%$: the weekly cycle explains the gain. Alone, it lowers the error of \cyc{} by 16.4\%, 42.5\% and
28.5\% on PEMS03, PEMS07 and PEMS08, more than \tsc{} does: \tsc{} has 3\%, 27\% and 14\% higher error than the trivial
competitor there, and 5.8\% lower error on Solar (Figure~\ref{fig:attr}b). On the development dataset ETTh1, validation picks the weekly cycle, which loses 2.9\%. The same document pre-registered a second test: re-ground the program on the base that the trivial
competitor selects and require a median gain of at least 3\% over that competitor. It fails, with a median change of
0.0\%. The re-grounded program stays within 1\% of the competitor on 7 of 8 datasets, and PEMS07 loses 4.3\%. Fitted in seconds on one CPU core, the trivial competitor is 12--41\% more accurate than the published CycleNet/Linear on all four PEMS datasets.

\subsection{Self-Improvement in Structure Space}\label{sec:res-struct}
\begin{figure*}[t]
\centering
\includegraphics[width=\linewidth]{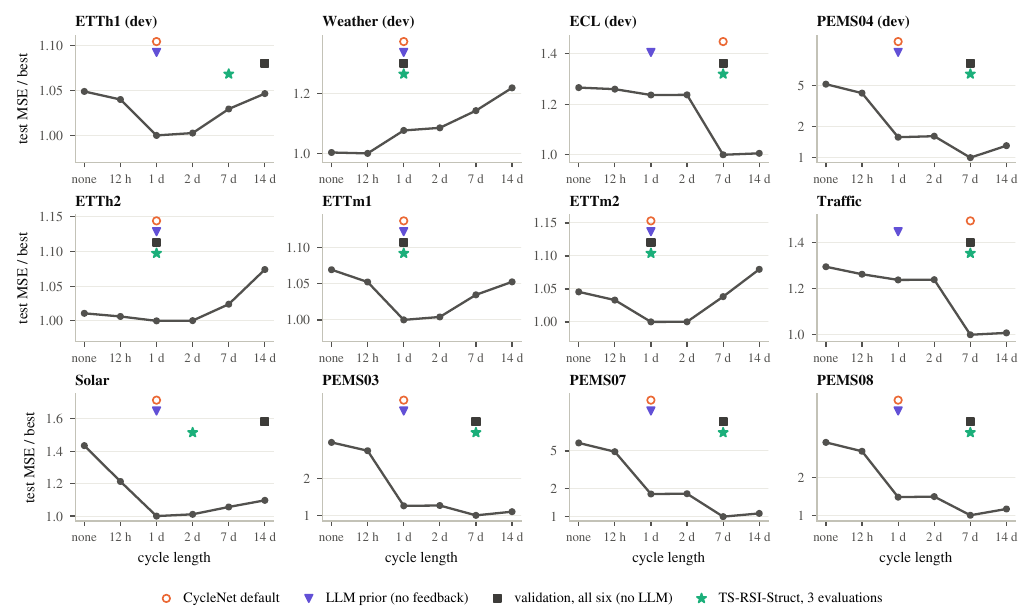}
\caption{Test MSE of \cyc{} for six cycle lengths, relative to the best (log scale on PEMS). Markers show each rule's
choice (\tss{}: most frequent over three seeds).}
\label{fig:period}
\end{figure*}

\subsubsection{The prior is wrong.} The programs the LLM writes compute a weekly profile, but asked for the cycle length
directly, the LLM does not supply this structure. Ranking the six cycle lengths from the dataset description, it puts one
day first on all 12 datasets, unanimously across samples (36 calls, 0.04~CNY). Its choice equals the CycleNet default except on ECL and Traffic, where it is 24\%
worse than the weekly default, and it is 9.8\% worse than six-way validation search at the median
(Figure~\ref{fig:period}). Random search needs three validation evaluations to match it at the median.

\subsubsection{The loop recovers from it.} The first proposal of \tss{} is always one day. After seeing the validation error
of that choice, the LLM proposes the weekly cycle next in 30 of 36 runs and third in the remaining six. After one round of
feedback ($k=2$) \tss{} matches six-way search at the median, and with two or three evaluations it is 4.7\% and 4.4\% more
accurate than random search over the same number of candidates. All three pre-registered criteria hold
(Figure~\ref{fig:struct}a). The static ranking truncated at two candidates also matches six-way search at the median over
the 12 datasets, but the loop pulls ahead on PEMS, where the static ranking puts the weekly cycle fourth. There, three
evaluations of \tss{} match exhaustive search, while the static ranking is 48\% and random search 14\% worse. After three evaluations the held-out geometric-mean change of \tss{}
relative to \cyc{} is $-12.4\%$, against $-12.5\%$ for the trivial competitor (Figure~\ref{fig:struct}b), and $-8.5\%$ after two.

\subsubsection{Other LLMs.} 
\begin{figure}[t]
\centering
\begin{subfigure}[t]{0.6\linewidth}\centering
\includegraphics[width=\linewidth]{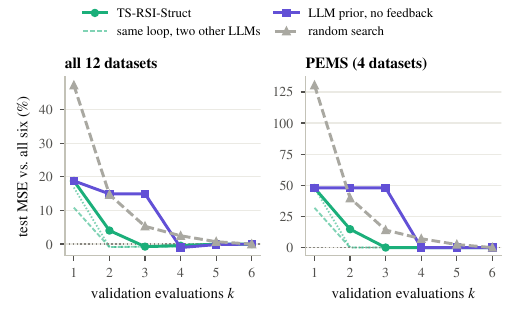}
\caption{Test MSE after $k$ evaluations.}
\end{subfigure}\hfill
\begin{subfigure}[t]{0.4\linewidth}\centering
\includegraphics[width=\linewidth]{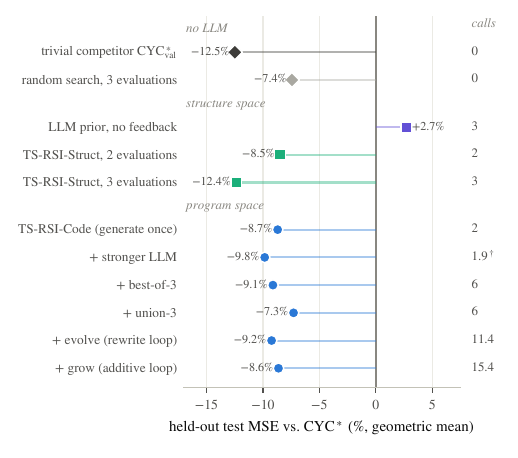}
\caption{All arms: gain and calls.}
\end{subfigure}
\caption{Self-improvement pays in structure space. (a) Test MSE relative to trying all six cycle lengths (mean over
datasets, whereas the criteria use medians). Thin lines: DeepSeek-V4-Pro (dashed) and Qwen3.6-Plus (dotted). (b) Held-out
geometric-mean change relative to \cyc{}. $^\dagger$Stronger LLM: about four times
the cost per call.}
\label{fig:struct}
\end{figure}

With the same prompts and criteria, DeepSeek-V4-Pro and Qwen3.6-Plus (another model family) both
pass all three criteria: with two and three evaluations they are 7.2\% and 4.4\% (DeepSeek-V4-Pro) and 7.2\% and 4.7\%
(Qwen3.6-Plus) more accurate than random search (Figure~\ref{fig:struct}a, thin lines), and both tie the trivial
competitor on the held-out datasets after three evaluations. Unlike DeepSeek-V4.1-Flash, both put the weekly cycle first
on ECL, so their static rankings are already as good at the median. Of their 216 answers each, 15 and 38 were invalid
and were replaced by the first untried candidate, as pre-registered.

\subsubsection{Two wrong choices.} 
\begin{figure*}[t]
\centering
\includegraphics[width=\linewidth]{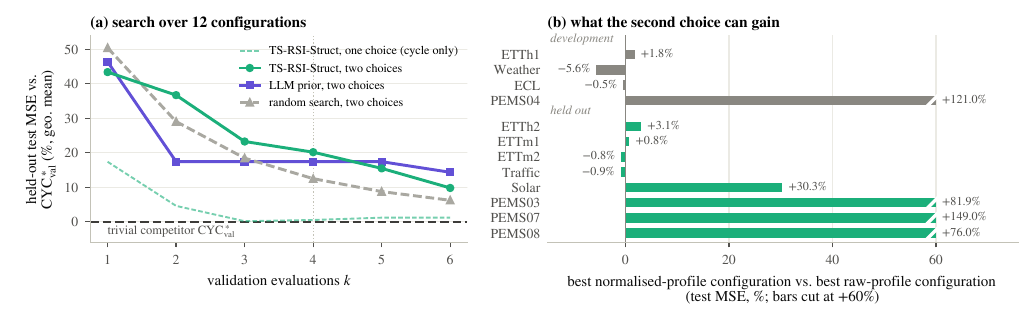}
\caption{Two structural choices: cycle length and profile scale (raw, as in \cyc{}, or normalised, as in CycleNet).
(a) Search over the 12 configurations (pre-registered comparison at $k=4$). (b) Best normalised vs.\ best raw
configuration, per dataset.}
\label{fig:struct2}
\end{figure*}

In a further
pre-registered experiment we added a second structural choice: whether the cycle profile is removed on the raw scale, as
in \cyc{}, or after instance normalisation, as in CycleNet, which gives 12 configurations per dataset. The second choice
leaves almost no room on the held-out datasets: the best normalised-profile configuration is at most 0.9\% better than
the best raw-profile one, and it is 30\% worse on Solar and 76--149\% worse on PEMS
(Figure~\ref{fig:struct2}b). The LLM's prior puts the daily cycle with the normalised profile first on all 12 datasets,
and in the loop it mostly varies the cycle length under that choice. In 3 of 12 PEMS runs it never tries the raw profile
in six evaluations. After four evaluations \tss{} is 20\% worse than the trivial competitor and 6.8\% worse than random
search, and all three criteria fail (Figure~\ref{fig:struct2}a).

\subsubsection{Cost.} Generating a program takes 1.75 calls on average (0.04~CNY) and grounding it a median of 14
CPU-seconds per dataset and horizon. The structure-space loop used 216 calls (0.24~CNY) for all 12 datasets and three
seeds. All experiments in this paper, including development runs, failed variants and 68 pilot calls, used 1867 calls
and 27.6~CNY (about 3.9 US dollars).

\section{Limitations}\label{sec:limits}
Our evidence is limited to LTSF benchmarks, which leave little room for domain knowledge beyond seasonality, and to
small structure spaces and short program loops. We did not separate the effects of proposing one candidate at a time and
of seeing validation scores. Finally, the held-out data were reused: the robust-grounding criteria were written after the
losses of CLG on ETTh2 and ETTm2 were known; evolve, best-of-3, union-3, the stronger LLM, the trivial competitor and
the ten-round loop after the held-out results of \tsc{}; \tss{} and its criteria after the test scores of all six cycle
lengths were known; and the human-written programs after we had seen LLM programs.

\section{Conclusion}
TS-RSI pairs every self-improvement loop on the LTSF benchmark with a trivial competitor. In program space, one grounded
LLM-written program captures the LLM's contribution, on PEMS a weekly cycle, and further rounds change the median by
about 1\% while inflating validation scores. In structure space, validation feedback moves the LLM off its prior within
one or two rounds. The loop beats random search with three LLMs and ties the trivial competitor for 0.24~CNY, but fails
when the prior is wrong on two choices. Self-improving forecasters should search small structure spaces with cheap,
informative feedback, and be evaluated against a trivial competitor for the structure they exploit, on a test set kept
out of selection.

\subsubsection*{Use of AI tools.} Apart from the LLMs studied in this paper, AI tools were used only to polish the
language of the manuscript. The authors reviewed all edited text and take full responsibility for the content.

%
%
%
\bibliographystyle{splncs04}
\bibliography{mybibliography}

\end{document}